\documentclass[10pt,twocolumn,letterpaper]{article}

\usepackage{cvpr}
\usepackage{pifont}
\usepackage[T1]{fontenc}
\usepackage[utf8]{inputenc}
\usepackage{amsmath}
\usepackage{amssymb}
\usepackage{graphicx}
\usepackage{booktabs}
\usepackage{multirow}
\usepackage{makecell}
\usepackage{xcolor}
\usepackage{colortbl}
\usepackage{marvosym}
\usepackage[accsupp]{axessibility} 

\definecolor{cvprblue}{rgb}{0.21,0.49,0.74}
\usepackage[pagebackref,breaklinks,colorlinks,allcolors=cvprblue]{hyperref}

\def\paperID{17275}
\def\confName{CVPR}
\def\confYear{2026}

\title{Learning Transferable Temporal Primitives\\ for Video Reasoning via Synthetic Videos}

\author{Songtao Jiang\textsuperscript{\rm 1}, Sibo Song\textsuperscript{\rm 2}, 
Chenyi Zhou\textsuperscript{\rm 1}, Yuan Wang\textsuperscript{\rm 1}, 
Ruizhe Chen\textsuperscript{\rm 1}, 
Tongkun Guan\textsuperscript{\rm 3},\\ Ruilin Luo\textsuperscript{\rm 4}, 
Yan Zhang\textsuperscript{\rm 1}, Zhihang Tang\textsuperscript{\rm 1}, 
Yuchong Sun\textsuperscript{\rm 2}, Hang Zhang\textsuperscript{\rm 2} \\
Zhibo Yang\textsuperscript{\rm 2}, Shuai Bai\textsuperscript{\rm 2}, 
Junyang Lin\textsuperscript{\rm 2}, 
Zuozhu Liu\textsuperscript{\rm 1}{\Letter}
\\[0.5em]
\textsuperscript{\rm 1} Zhejiang University \quad
\textsuperscript{\rm 2} Qwen Team, Alibaba Group \\
\textsuperscript{\rm 3} Shanghai Jiao Tong University \quad
\textsuperscript{\rm 4} Tsinghua University \\
{\tt\small zuozhu.liu@zju.edu.cn}
}

\begin{document}
\maketitle

\begin{abstract}
The transition from image to video understanding requires vision-language models (VLMs) to shift from recognizing static patterns to reasoning over temporal dynamics such as motion trajectories, speed changes, and state transitions. Yet current post-training methods fall short due to two critical limitations: (1) existing datasets often lack temporal-centricity, where answers can be inferred from isolated keyframes rather than requiring holistic temporal integration; (2) training data generated by proprietary models contains systematic errors in fundamental temporal perception, such as confusing motion directions or misjudging speeds. We introduce \textbf{SynRL}, a post-training framework that teaches models \textit{temporal primitives}, the fundamental building blocks of temporal understanding including direction, speed, and state tracking. Our key insight is that these abstract primitives, learned from programmatically generated synthetic videos, transfer effectively to real-world scenarios. We decompose temporal understanding into short-term perceptual primitives (speed, direction) and long-term cognitive primitives (state tracking, retrodictive inference), constructing 7.7K CoT and 7K RL samples with ground-truth frame-level annotations through code-based video generation. Despite training on simple geometric shapes, SynRL achieves substantial improvements across 15 benchmarks spanning temporal grounding, complex reasoning, and general video understanding. Remarkably, our 7.7K synthetic CoT samples outperform Video-R1's 165K real-world samples. We attribute this to fundamental temporal skills—such as tracking frame-by-frame changes and comparing velocity—that transfer effectively from abstract synthetic patterns to complex real-world scenarios. This establishes a new paradigm for video post-training: video temporal learning through carefully designed synthetic data provides a more cost-efficient scaling path. Code is released on \href{https://github.com/jiangsongtao/Synthetic-Video}{https://github.com/jiangsongtao/Synthetic-Video}.

\end{abstract}

\begin{figure}[t]
    \centering
    \includegraphics[width=0.8\linewidth]{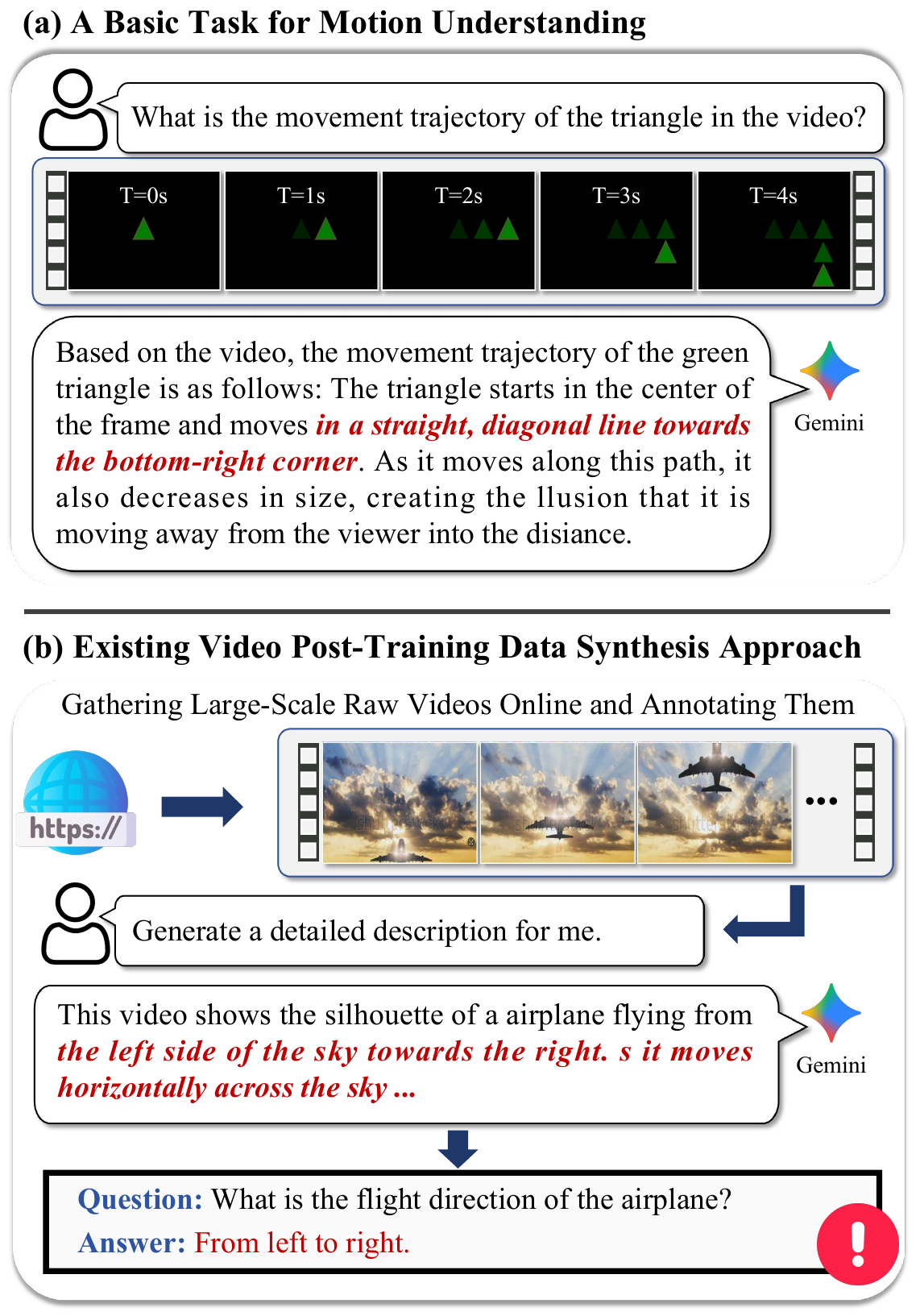}
    \caption{\textbf{Systematic failures in proprietary models' temporal perception.} 
(\textbf{a}) Gemini-2.5-Pro incorrectly describes a simple geometric shape's 
motion trajectory. (\textbf{b}) When used to annotate real-world videos, such 
flawed descriptions propagate errors into training data. 
(\textcolor[RGB]{195,0,0}{Erroneous phrases in red.})}
    \label{fig:motivation}
\end{figure}
    
\section{Introduction}

The evolution of vision-language models (VLMs) from static image analysis to 
dynamic video understanding necessitates a fundamental shift in core capabilities. 
While image models excel at recognizing static visual patterns such as objects and
scenes, video understanding requires reasoning over 
temporal dynamics, encompassing motion trajectories, speed changes, and 
sequential state transitions. This transition demands that models develop 
\textit{temporal primitives}: fundamental building blocks of temporal perception 
including direction, speed, acceleration, and state tracking.
Reinforcement learning (RL)-based post-training has emerged as a promising 
paradigm for enhancing these capabilities~\cite{feng2025videor1reinforcingvideoreasoning,li2025videochatr1enhancingspatiotemporalperception}. 
However, the scarcity of high-quality temporally-annotated video data forces 
current approaches to rely on proprietary models (e.g., GPT-4V, Gemini-2.5-pro~\cite{Gemini2.5}) to 
synthesize training data, either generating qa pairs or producing 
chain-of-thought (CoT) annotations~\cite{zhang2024video,wang2025jigsaw,kong2025tuna}. This dependency introduces a critical 
bottleneck: even proprietary models exhibit systematic failures 
in fundamental temporal reasoning.

\begin{figure*}[t]
    \centering
    \includegraphics[width=0.95\linewidth]{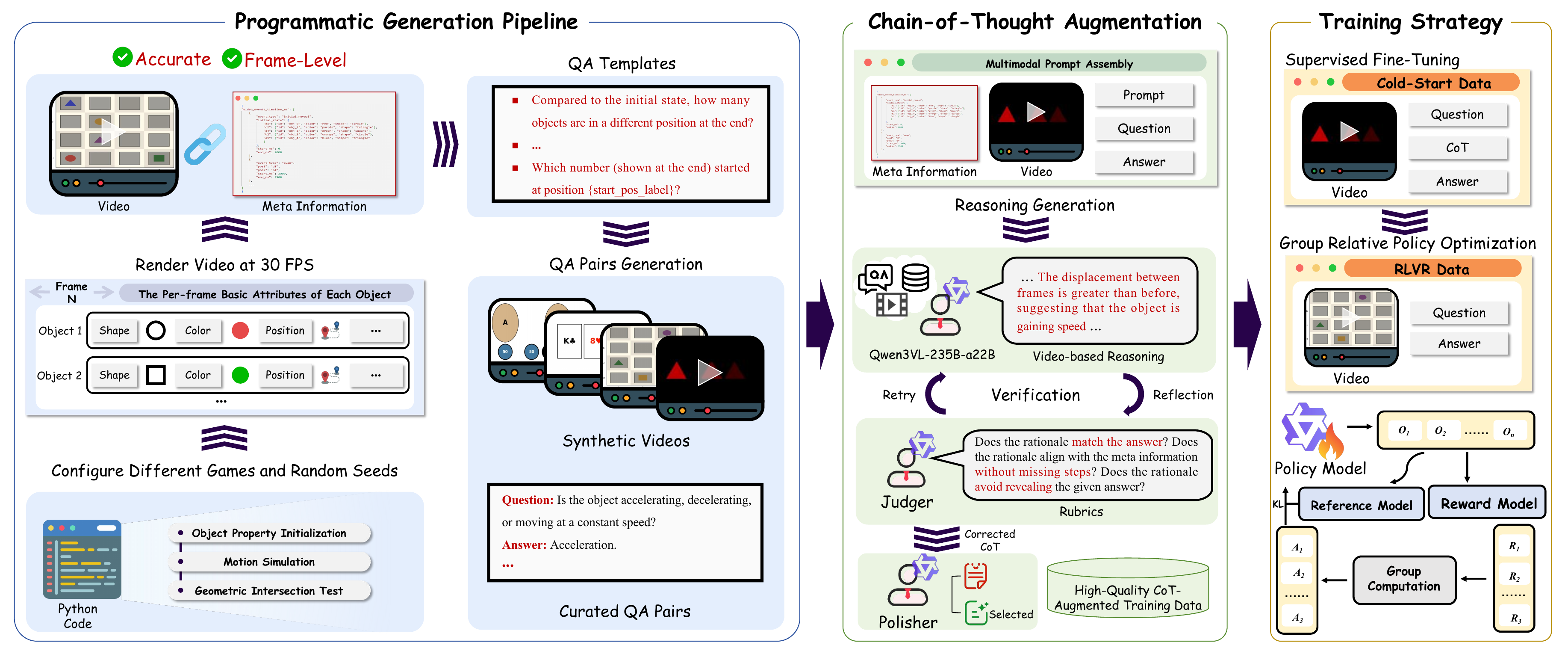}
    \caption{\textbf{Overview of SynRL framework.} 
(\textit{Left}) \textbf{Programmatic Generation Pipeline:} Object properties 
and motion dynamics are specified in Python code to generate videos with 
accurate frame-level metadata. Videos are rendered at 30 FPS, and QA pairs are 
instantiated from hand-crafted templates conditioned on the same metadata, 
yielding temporally grounded synthetic video-QA triples. 
(\textit{Middle}) \textbf{Chain-of-Thought Augmentation:} Given a synthetic 
video, its metadata, and its QA pair, a multimodal LLM generates step-by-step 
reasoning chains. A Judger verifies their consistency with the event timeline, 
filtering out incorrect outputs. The verified reasoning chains are then passed 
to a Polisher, which refines them into more natural and fluent CoT annotations 
while preserving factual correctness. 
(\textit{Right}) \textbf{Training Strategy:} The target model is first trained 
via supervised fine-tuning on the polished CoT data to learn explicit temporal 
reasoning patterns. It is then optimized with group relative policy optimization 
(GRPO) on synthetic video-QA samples with verifiable rewards, so that 
reinforcement learning further improves reasoning quality under strictly correct 
supervisory signals.}
    \label{fig:pipeline}
\end{figure*}
Figure~\ref{fig:motivation} illustrates this brittleness. When prompted to 
describe basic temporal events, Gemini-2.5-Pro produces fluent but factually 
incorrect explanations. In Figure~\ref{fig:motivation}(a), it misidentifies a 
simple geometric shape's motion trajectory. Figure~\ref{fig:motivation}(b) 
shows similar failures on natural videos, where the model hallucinates motion 
direction. When such flawed descriptions are used to construct training data 
(Figure~\ref{fig:motivation}(b)), the resulting QA pairs or CoT 
annotations inherit these systematic errors, teaching models a 
``fluent-but-wrong'' reasoning style that poisons the supervisory signal.
Beyond data quality, existing video datasets suffer from a second fundamental 
limitation: lack of \textit{temporal-centricity}. Many contain questions 
answerable from isolated keyframes, allowing models to bypass true temporal 
integration through static pattern matching. This raises a pivotal question: 
 \textit{How can we obtain high-quality, 
temporal-centric training data without relying on proprietary models?}

To address these challenges, we introduce \textbf{SynRL}, a post-training framework 
that teaches temporal primitives through programmatically generated synthetic 
videos with guaranteed ground-truth annotations. \textbf{\textit{Our key insight is that these 
abstract primitives, while learned from simple synthetic scenarios, transfer 
effectively to complex real-world videos.}} By generating videos through code, 
we obtain frame-level procedural metadata including state snapshots, event 
timestamps, and operation sequences, enabling precise temporal annotation while 
bypassing proprietary models' flawed perception.
Crucially, our data is {temporal-centric by design}. Questions are 
constructed to be unanswerable from isolated keyframes, requiring models to 
track dynamic events, integrate information across sequences, and reason about 
temporal changes. We decompose temporal understanding into two categories: 
(1) \textit{short-term perceptual primitives} testing fundamental motion 
perception (direction, speed, motion) over brief time windows, and 
(2) \textit{long-term perceptual primitives} evaluating sustained reasoning 
(state tracking, retrodictive inference) over extended sequences. Through 
this pipeline, we generate 6.7K CoT samples and 7K RL samples. To prevent 
distribution shift during cold-start, we supplement with 1K real-world 
video-QA pairs from LLaVA-Video~\cite{zhang2025llavavideovideoinstructiontuning}, included {without} 
their original CoT to avoid incorrect reasoning chains.
Our training follows a two-stage process. First, we perform supervised 
fine-tuning (SFT) on 7.7K mixed CoT samples, teaching the structure of correct temporal reasoning. Second, 
we conduct Group Relative Policy Optimization (GRPO)~\cite{shao2024deepseekmath} using 7K 
synthetic video-QA pairs, where the model improves reasoning correctness 
through RL guided by verifiable accuracy rewards.

Despite training on stark simplicity, SynRL demonstrates remarkable generalization. 
Fundamental skills learned from synthetic videos transfer directly to natural 
videos involving human actions, camera motion, and complex scene dynamics. 
Extensive evaluation across 15 benchmarks reveals consistent improvements: 
+12.6\% on RexTime~\cite{chen2024rextime} for temporal grounding and +4.6\% on TOMATO~\cite{shangguan2024tomato} for complex 
reasoning. Strikingly, our 7.7K synthetic CoT samples achieve superior results 
compared to Video-R1's 165K real-world videos CoT samples~\cite{feng2025videor1reinforcingvideoreasoning}, representing 
a 21× improvement in data efficiency.
{Our contributions are:}
\begin{itemize}[leftmargin=*,noitemsep,topsep=2pt]
\item We identify systematic temporal perception errors in proprietary models 
that poison current video post-training paradigms (Figure~\ref{fig:motivation}).

\item We introduce SynRL, a framework generating temporal-centric synthetic 
videos with guaranteed ground-truth annotations, decomposing temporal 
understanding into learnable primitives.

\item We demonstrate that abstract primitives learned from simple synthetic 
scenarios transfer to complex real-world videos, achieving substantial gains 
across 15 benchmarks with 21× greater data efficiency, establishing a new 
paradigm for video post-training.
\end{itemize}

\begin{figure*}[t!]
    \centering
    \includegraphics[width=0.9\linewidth]{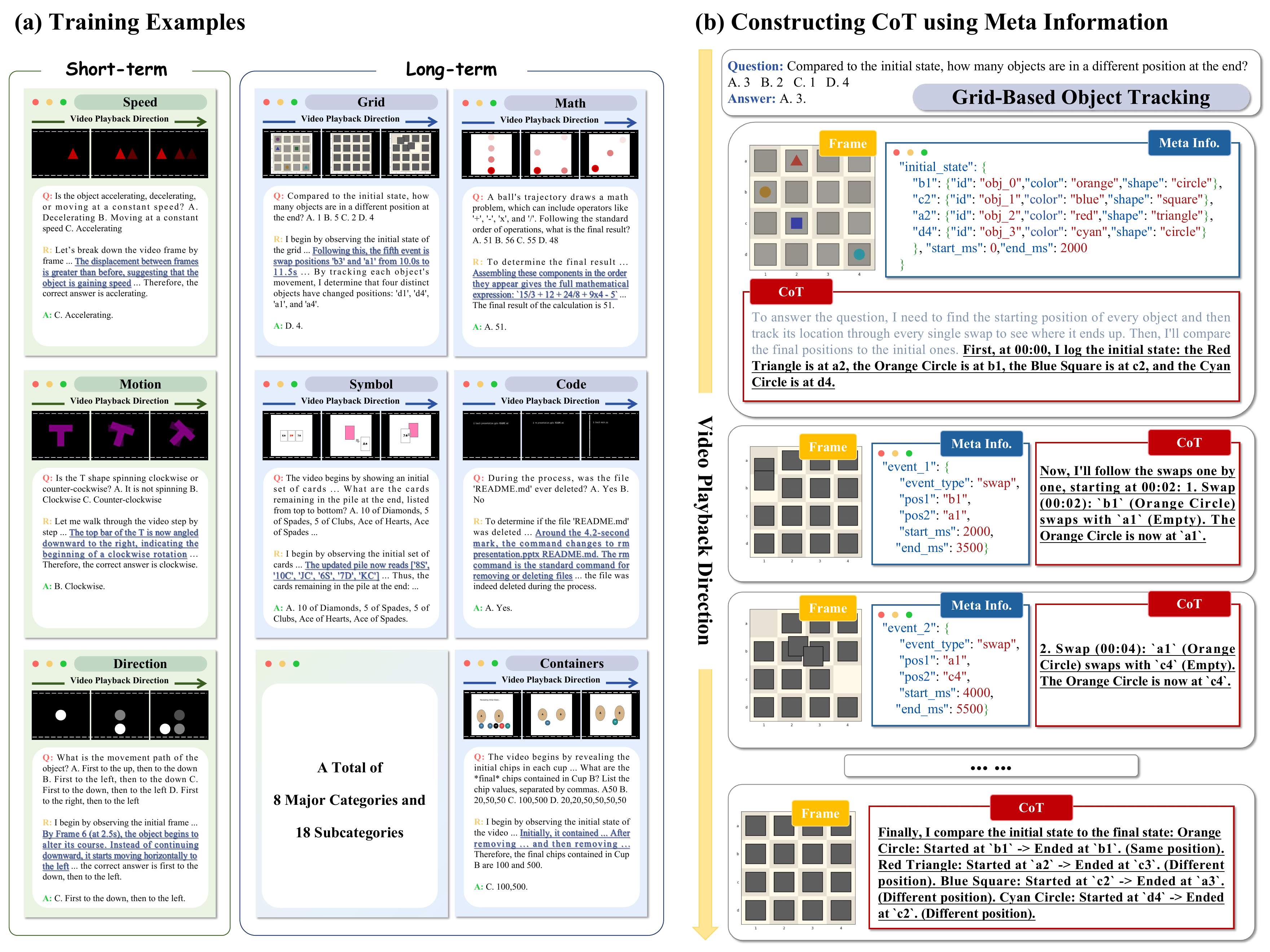}
    \caption{\textbf{(a) Training Examples:} We generate diverse synthetic videos spanning 8 major categories with 18 subcategories, covering short-term perceptual primitives (speed perception, motion tracking, direction identification) and long-term cognitive primitives (grid-based object tracking, symbol manipulation, code execution, mathematical operations, container management). Each video is procedurally generated with verifiable ground-truth answers.
\textbf{(b) Constructing CoT using Meta Information:} Using a grid-based object tracking game as an example, we provide the video with its code-derived metadata (initial state, swap events, timestamps) to a VLM. The model generates step-by-step reasoning chains that track each object through every swap operation, explicitly referencing frame timestamps (e.g., ``at 00:00'', ``at 00:02'') and logging state transitions. This metadata-conditioned generation ensures temporal grounding and verifiable correctness, as the reasoning must align with the documented event sequence.
}
    \label{fig:rationale}
\end{figure*}


\section{Methods}
\label{sec:methods}
As illustrated 
in Figure~\ref{fig:pipeline}, our approach consists of three stages: 
(1) \textit{Programmatic Generation} (\S\ref{subsec:generation}), where we 
produce synthetic videos with guaranteed ground-truth annotations covering 
short-term perceptual and long-term cognitive primitives; 
(2) \textit{Chain-of-Thought Augmentation} (\S\ref{subsec:cot_data}), where 
we construct high-quality reasoning annotations through iterative verification 
and polishing; and (3) \textit{Training Strategy} (\S\ref{subsec:training}), 
where we adopt a two-stage approach combining supervised fine-tuning (SFT) on CoT 
data and RL with verifiable rewards.

\subsection{Programmatic Generation Pipeline}
\label{subsec:generation}
\subsubsection{Short-Term Motion Videos}
\label{subsec:short_term}
Short-term videos test fundamental perceptual abilities over brief time windows, 
presenting simple, abstract motion patterns where models must identify directions, 
recognize trajectory shapes, detect acceleration, or count discrete events. The 
core challenge lies in attending to frame-by-frame changes rather than relying 
on static visual pattern recognition.
We implement 12 synthetic video types covering diverse perceptual skills: 
\textit{collision event counting} (tracking wall contacts during continuous 
motion), \textit{direction identification} (determining movement orientation), 
\textit{trajectory shape recognition} (classifying motion paths as linear, 
circular, or zigzag), \textit{speed perception} (comparing relative velocities), 
\textit{motion counting} (tallying discrete movement events), \textit{attribute 
change detection} (identifying color, size, or shape transformations), 
\textit{rotation perception} (counting full rotations), \textit{relative 
position tracking} (monitoring spatial relationships), \textit{acceleration 
detection} (recognizing speed changes), \textit{velocity comparison} (ranking 
objects by speed), \textit{distance estimation} (judging spatial distances between moving objects), and \textit{sequential event ordering} (determining temporal 
order of discrete actions). 

\noindent\textbf{Generation Pipeline.}
We initialize object properties (shape, color, size, position, velocity) and simulate motion using basic physics equations. The simulation iteratively updates positions based on velocity, detects boundary collisions through geometric intersection tests, and increments rotation angles for spinning objects. We track key events such as wall contacts or direction changes with frame-level timestamps. Ground-truth answers are computed directly from simulation logs: collision counts from event counters, trajectory shapes from position trace analysis, and rotation counts from cumulative angular displacement. Videos are rendered at 30 FPS using Python's Matplotlib library and encoded to MP4 format with H.264 for efficient storage.

\subsubsection{Long-Term Perception Videos}
\label{subsec:long_term}

While short-term videos test immediate perception, long-term videos evaluate 
sustained reasoning over extended sequences. These videos depict scenarios 
where an initial state undergoes a series of discrete operations, with the 
state visible only at the beginning or end—never during transitions. Following recent work that employs long-term synthetic videos to evaluate the reasoning capabilities of video language models~\cite{liu2025videoreasonbench,NEURIPS2024_329ad516}, we investigate whether such a construction paradigm can also benefit models during the training phase. Specifically, we conceptualize synthetic videos as state transition sequences $\{S_t, o_t, S_{t+1}\}$, where each operation $o_t$ transforms state $S_t$ into the subsequent state $S_{t+1}$. The operation 
sequence remains observable, but intermediate states remain latent. This 
design forces models to maintain internal representations and perform 
step-by-step reasoning to infer hidden configurations.
We implement 6 scenario types spanning: (1) \textit{abstract data structure 
tracking}, including card stack manipulation, chip container management, file 
system navigation, and mathematical symbol operations; (2) \textit{grid-based 
object tracking} with visual occlusion, such as shell games where objects are 
shuffled behind barriers; and (3) \textit{retrodictive identity inference}, 
such as sliding number puzzles requiring backward reasoning. All scenarios 
follow a unified structure: an \textit{initial state} $S_0$ is shown, followed 
by a sequence of \textit{state-changing operations} $\{o_1, o_2, \ldots, o_T\}$ 
that produces a \textit{final state} $S_T$. Critically, only one of the two 
states is visible either $S_0$ or $S_T$, while the operation sequence remains 
fully observable. This enforces a reasoning challenge: models cannot simply 
compare visible states but must internally simulate the complete state evolution.
Questions target different reasoning modes: forward prediction (inferring $S_T$ 
from $S_0$ and operations), retrodictive inference (inferring $S_0$ from $S_T$ 
by mentally reversing the sequence), and historical queries (reconstructing 
intermediate states or operation properties).

\noindent\textbf{Generation Pipeline.}
We generate videos programmatically using Python's Matplotlib library. Each 
video consists of three phases: (1) \textit{initial reveal}, displaying $S_0$ 
or a blank screen for 2 seconds; (2) \textit{operation sequence}, showing 
smooth animated transitions for each operation $o_t$ (0.5-1 second per 
operation); and (3) \textit{final reveal}, displaying $S_T$ or a blank screen 
for 2 seconds. We maintain complete state histories with frame-indexed metadata 
recording each operation's timing and effects. For multiple-choice questions, 
we automatically generate distractors by applying random alternative operations 
to the correct state. Videos are composited and encoded using FFmpeg with 
consistent frame rate and resolution. 

\subsection{Chain-of-Thought Augmentation}
\label{subsec:cot_data}
To train models to generate explicit reasoning, we construct CoT 
annotations using a novel four-stage pipeline that leverages our procedural 
metadata—frame-indexed events with millisecond timestamps, state snapshots, 
and complete operation sequences (as shown in Figure~\ref{fig:rationale}). We regenerate videos with distinct content 
and non-overlapping questions from the RL data to prevent overfitting.
\noindent\textbf{1. Generation.}
We prompt a multimodal model (Qwen3-VL-235B-A22B~\cite{bai2025qwen3vltechnicalreport}) with the synthesized video, the 
corresponding question and ground-truth answer, and the complete event timeline 
as reference. The model generates step-by-step reasoning that simulates a 
person watching the video and arriving at the correct answer through explicit 
temporal analysis.
\noindent\textbf{2. Verification.}
Since models can produce flawed reasoning inconsistent with ground-truth 
annotations, we employ a separate LLM evaluator (Qwen3-235B-A22B) to validate each 
reasoning chain. The evaluator checks whether the generated CoT: (1) reaches 
the correct final answer, and (2) accurately follows the event sequence with 
proper temporal alignment to the annotated metadata.
\noindent\textbf{3. Reflection.}
If verification fails, the evaluator generates detailed feedback explaining the 
inconsistency, which is sent back to the generator for refinement. We allow up 
to 5 iterations; reasoning chains that fail validation after this limit are 
filtered out. This iterative process substantially improves reasoning quality 
and correctness.
\noindent\textbf{4. Polishing.}
Once verified, we use an LLM (Qwen3-235B-A22B) to refine the reasoning chain, making 
the language more natural and fluent while preserving all logical content and 
temporal dependencies.

\subsection{Training Strategy}
\label{subsec:training}

We adopt a two-stage training approach optimized for data efficiency and 
reasoning quality.
\noindent\textbf{Stage 1: Supervised Fine-Tuning.}
We perform SFT using our generated CoT data to teach models step-by-step 
temporal reasoning. To prevent distribution shift and preserve general video 
understanding capabilities, we adopt a mixed curriculum strategy, combining 
synthetic temporal reasoning videos with general video QA data from 
LLaVA-Video~\cite{zhang2025llavavideovideoinstructiontuning} (approximately 15\% of the training mixture). 
Critically, synthetic videos receive full CoT supervision generated via our 
pipeline, while general videos use only direct answer supervision without CoT. 
This prevents error propagation from potentially flawed model-generated reasoning 
on general data, while ensuring models learn temporal reasoning from high-quality 
supervision on synthetic data. 
\noindent\textbf{Stage 2: Reinforcement Learning with GRPO.}
Since models already learned to generate structured reasoning during SFT, RL 
focuses on improving reasoning correctness rather than teaching reasoning format. 
We conduct GRPO~\cite{shao2024deepseekmath} using only 
accuracy rewards on our synthetic videos. We construct approximately {7K 
RL samples} covering both short-term and long-term scenarios. The verifiable 
nature of our synthetic data enables perfectly correct reward signals, allowing 
the model to refine its reasoning under strictly reliable supervision.

\newcommand{\showdelta}[2]{%
  \begin{tabular}{@{}c@{}}%
    #1 \\
    {\small\textcolor{red}{+#2}}%
  \end{tabular}%
}
\definecolor{gred}{HTML}{cc0200}
\definecolor{ggreen}{HTML}{4C9F26}
\newcommand{\daa}[1]{\scriptsize\textcolor{ggreen}{\footnotesize $\downarrow$}{\color{ggreen}#1}}
\newcommand{\uaa}[1]{\scriptsize\textcolor{gred}{\footnotesize $\uparrow$}{\color{gred}#1}}
\begin{table*}[htbp]
\centering
  \caption{Performance comparison on diverse video benchmarks. }
\label{tab:main_results}
\resizebox{0.8\textwidth}{!}{%
\begin{tabular}{l cc cccc}
\toprule
& \multicolumn{2}{c}{\textbf{Complex Reasoning}} & \multicolumn{4}{c}{\textbf{Comprehensive Video Understanding}} \\
\cmidrule(lr){2-3} \cmidrule(lr){4-7}
\textbf{Models} & \textbf{TOMATO} & \textbf{Video-TT} & \textbf{MVBench} & \textbf{VideoMME} & \multicolumn{2}{c}{\textbf{TemporalBench}} \\
\cmidrule(lr){6-7}
& & & & & \small Binary Acc. & \small Multiple Binary Acc. \\
\midrule
\rowcolor[HTML]{FFFFFF} LLaVA-OneVision-7B & 25.5 & - & 58.6 & 63.3 & 61.9 & 21.2 \\
\rowcolor[HTML]{FFFFFF} NVILA-8B & - & - & 68.1 & 64.2 & - & - \\
\rowcolor[HTML]{FFFFFF} Apollo-7B & - & - & - & 61.3 & - & - \\
\rowcolor[HTML]{FFFFFF} VideoLLaMA-2-7B & 10.1 & - & 57.3 & 54.9 & - & - \\
\rowcolor[HTML]{FFFFFF} VideoLLaMA3-7B & 30.1 & - & 69.7 & 66.2 & 69.1 & 29.4 \\
\rowcolor[HTML]{FFFFFF} Video-R1-7B & 25.1 & - & 63.9 & 59.3 & 63.7 & 21.2 \\
\rowcolor[HTML]{FFFFFF} Video-Jigsaw-7B & 25.3 & - & 53.1 & - & 60.8 & 20.4 \\
\midrule
\rowcolor[HTML]{FFFFEA} Qwen3-VL-4B & 32.1 & 38.9 & 65.4 & 60.9 & 67.4 & 27.1 \\
\rowcolor[HTML]{D7F9F8} Qwen3-VL-4B + SynRL & 36.7 \uaa{4.6} & 40.7 \uaa{1.8} & 67.1 \uaa{1.7} & 62.0 \uaa{1.1} & 68.2 \uaa{0.8} & 27.9 \uaa{0.8} \\
\rowcolor[HTML]{FFFFEA} Qwen3-VL-8B & 33.2 & 40.6 & 67.2 & 63.4 & 68.3 & 28.5 \\
\rowcolor[HTML]{D7F9F8} Qwen3-VL-8B + SynRL & 38.1 \uaa{4.9} & 41.5 \uaa{0.9} & 69.1 \uaa{1.9} & 65.2 \uaa{1.8} & 69.3 \uaa{1.0} & 29.4 \uaa{0.9} \\
\bottomrule
\end{tabular}%
}
\end{table*}

\begin{table*}[htbp]
\centering
\caption{Performance comparison on additional video benchmarks.}
\label{tab:main_results2}
\resizebox{0.8\textwidth}{!}{%
\begin{tabular}{l cccc}
\toprule
\textbf{Benchmark} & \textbf{Qwen3-VL-4B} & \textbf{Qwen3-VL-4B + SynRL} & \textbf{Qwen3-VL-8B} & \textbf{Qwen3-VL-8B + SynRL} \\
\midrule
\multicolumn{5}{l}{\textit{General Video QA}} \\
\midrule
Aotbench & \cellcolor[HTML]{FFFFEA} 52.7 & \cellcolor[HTML]{D7F9F8} 54.4 \uaa{1.7} & \cellcolor[HTML]{FFFFEA} 54.8 & \cellcolor[HTML]{D7F9F8} 57.7 \uaa{2.9} \\
vinoground & \cellcolor[HTML]{FFFFEA} 40.8 & \cellcolor[HTML]{D7F9F8} 43.2 \uaa{2.4} & \cellcolor[HTML]{FFFFEA} 43.4 & \cellcolor[HTML]{D7F9F8} 47.6 \uaa{4.2} \\
VideoMMMU & \cellcolor[HTML]{FFFFEA} 54.1 & \cellcolor[HTML]{D7F9F8} 54.5 \uaa{0.4} & \cellcolor[HTML]{FFFFEA} 60.0 & \cellcolor[HTML]{D7F9F8} 61.0 \uaa{1.0} \\
\midrule
\multicolumn{5}{l}{\textit{Complex Reasoning}} \\
\midrule
vcrbench & \cellcolor[HTML]{FFFFEA} 30.1 & \cellcolor[HTML]{D7F9F8} 31.5 \uaa{1.4} & \cellcolor[HTML]{FFFFEA} 34.2 & \cellcolor[HTML]{D7F9F8} 35.6 \uaa{1.4} \\
cvbench & \cellcolor[HTML]{FFFFEA} 54.1 & \cellcolor[HTML]{D7F9F8} 54.9 \uaa{0.8} & \cellcolor[HTML]{FFFFEA} 53.8 & \cellcolor[HTML]{D7F9F8} 54.9 \uaa{1.1} \\
TVbench & \cellcolor[HTML]{FFFFEA} 51.9 & \cellcolor[HTML]{D7F9F8} 53.6 \uaa{1.7} & \cellcolor[HTML]{FFFFEA} 54.0 & \cellcolor[HTML]{D7F9F8} 54.7 \uaa{0.7} \\
Minerva & \cellcolor[HTML]{FFFFEA} 32.5 & \cellcolor[HTML]{D7F9F8} 32.8 \uaa{0.3} & \cellcolor[HTML]{FFFFEA} 34.4 & \cellcolor[HTML]{D7F9F8} 35.2 \uaa{0.8} \\
\bottomrule
\end{tabular}%
}
\end{table*}

\begin{table*}[htbp]
\centering
\caption{Performance comparison on NExTGQA and RexTime benchmarks. It can be observed that SynRL outperforms existing SFT-based methods trained with large-scale data.}
\label{tab:grounding_results}
\begin{tabular}{l|cccc|cccc}
\toprule
\textbf{Model} & \multicolumn{4}{c|}{\textbf{NExTGQA}} & \multicolumn{4}{c}{\textbf{RexTime}} \\
               & \textbf{R@0.3} & \textbf{R@0.5} & \textbf{R@0.7} & \textbf{mIoU} & \textbf{R@0.3} & \textbf{R@0.5} & \textbf{R@0.7} & \textbf{mIoU} \\
\midrule
Qwen2.5-VL-7B          & 31.6 & 18.1 & 7.5 & 20.9 & 10.3 & 6.1 & 3.0 & 8.1 \\
Qwen2.5-VL-32B         & 38.0 & 22.3 & 10.0 & 25.4 & 16.8 & 10.0 & 5.1 & 13.0 \\
VTimeLLM               & 37.9 & 20.2 & 9.7 & 24.4 & 28.8 & 17.4 & 7.2 & 20.1 \\
TimeChat               & 34.1 & 17.9 & 6.2 & 20.6 & 14.4 & 7.6 & 3.1 & 11.7 \\
VideoChat-TPO          & 41.2 & 23.4 & 8.2 & 27.7 & 34.5 & 19.3 & 9.8 & 25.2 \\
\midrule
\rowcolor[HTML]{FFFFEA}Qwen3-VL-4B & 32.9 & 20.0 & 9.8 & 23.5 & 26.2 & 20.1 & 12.9 & 20.9 \\
\rowcolor[HTML]{FFFFEA}Qwen3-VL-4B CoT & 33.7 & 20.2 & 9.4 & 23.7 & 25.4 & 18.0 & 12.6 & 20.1 \\
\rowcolor[HTML]{D7F9F8}Qwen3-VL-4B + SynRL & 38.8 \uaa{5.9} & 21.8 \uaa{1.8} & 10.2 \uaa{0.4} & 28.1 \uaa{4.6} & 38.8 \uaa{12.6} & 26.2 \uaa{6.1} & 16.3 \uaa{3.4} & 28.9 \uaa{8.0} \\
\bottomrule
\end{tabular}
\end{table*}

\begin{table}[htbp]
\centering
\caption{Performance comparison on Charades dataset}
\label{tab:charades_performance}
\resizebox{0.5\textwidth}{!}{
\begin{tabular}{lrrrr}
\toprule
\textbf{Model} & \textbf{R@0.3} & \textbf{R@0.5} & \textbf{R@0.7} & \textbf{mIoU} \\
\midrule
VTimeLLM & 55.3 & 34.3 & 14.7 & 34.6 \\
TimeChat & 51.5 & 32.2 & 13.4 & - \\
VideoChat-TPO & 58.3 & 40.2 & 18.4 & 38.1 \\
\midrule
\rowcolor[HTML]{FFFFEA}Qwen3-VL-4B & 65.1 & 41.6 & 20.5 & 41.9 \\
\rowcolor[HTML]{FFFFEA}Qwen3-VL-4B CoT & 53.6 & 33.2 & 14.7 & 34.5 \\
\rowcolor[HTML]{D7F9F8}Qwen3-VL-4B + SynRL & 73.7 \uaa{8.6} & 50.1 \uaa{8.5} & 23.0 \uaa{2.5} & 47.0 \uaa{5.1} \\
\bottomrule
\end{tabular}
}
\end{table}

\section{Experiments}
\label{sec:expe}

\subsection{Experimental Setup}
We evaluate our model on a comprehensive suite of 15 video benchmarks spanning three categories: \textbf{(1) Temporal Grounding:} We assess temporal localization capabilities on NExTGQA~\cite{xiao2024can}, Charades-STA~\cite{gao2017tall}, and RexTime~\cite{chen2024rextime}, which require models to accurately predict timestamps of specific actions or events. Following standard protocols, we report R@0.3, R@0.5, R@0.7 (recall at different IoU thresholds), mean Intersection over Union (mIoU) for measuring the overlap between predicted and ground-truth segments, and mean Intersection over Prediction (mIoP) for evaluating prediction precision. \textbf{(2) Complex Reasoning:} We evaluate on TOMATO~\cite{shangguan2024tomato}, a benchmark designed to assess temporal reasoning across human-centric and real-world scenarios; TemporalBench~\cite{cai2024temporalbench}, which evaluates fine-grained temporal dynamics including action frequency and motion magnitude; Video-TT~\cite{zhang2025towards} for temporal understanding; and complex reasoning benchmarks VCRBench~\cite{sarkar2025vcrbench}, CVBench~\cite{zhu2025cvbench}, TVBench~\cite{cores2024tvbench}, and Minerva~\cite{Nagrani_2025_ICCV} that challenge models on causal relationships and cross-video understanding. \textbf{(3) General Video Understanding:} To verify generalizability, we evaluate on Video-MME~\cite{fu2025video} and MVBench~\cite{li2024mvbench} for comprehensive video understanding across diverse video types and durations, as well as VideoMMMU~\cite{hu2025video}, AoTBench~\cite{aot}, and vinoground~\cite{zhang2024vinoground} for multimodal understanding and compositional reasoning. For all benchmarks in categories (2) and (3), we report accuracy as the evaluation metric, consistent with prior work.

\subsection{Baselines}

We compare our approach against three categories of baselines:
\noindent\textbf{(1) Pretrained VLMs:} We include representative vision-language models as reference points, such as LLaVA-OneVision-7B~\cite{li2024llava} and the VideoLLaMA~\cite{cheng2024videollama} series. These models achieve comprehensive video understanding through large-scale SFT on diverse multimodal datasets.
\noindent\textbf{(2) RL-based Methods:} We compare with recent reinforcement learning approaches including Video-R1~\cite{feng2025videor1reinforcingvideoreasoning} and Video-Jigsaw~\cite{Wu2025VisualJP}. Video-R1 employs a two-stage training pipeline: cold-start SFT on Video-R1-CoT-165k followed by RL training on Video-R1-260k, both comprising mixed image and video data. Video-Jigsaw performs RL training on 100K shuffled videos from the LLaVA-Video dataset to enhance temporal understanding.
\noindent\textbf{(3) Base Models:} We adopt Qwen2.5-VL-7B as our primary base model~\cite{Qwen2.5VL}, which has been widely adopted in recent RL-based vision-language research, and additionally experiment with Qwen3-VL-4B, the state-of-the-art model within the 7B parameter range. For both base models, we train using our synthesized dataset consisting of 5K chain-of-thought annotated video samples for SFT cold-start and 5K samples for subsequent RL training.

\subsection{Implementation Details}
All experiments are conducted on 8 NVIDIA H100 GPUs (80GB). We employ the VeRL framework to implement the GRPO algorithm. Specifically, we remove both the KL regularization term and the entropy loss to enable more aggressive policy updates. Training is performed with a global batch size of 512 and a learning rate of $1 \times 10^{-6}$. For each prompt during training, we sample 8 candidate responses using nucleus sampling with a temperature of 1.0 to ensure sufficient exploration.

\subsection{Results}

\noindent\textbf{Strong Generalization of Synthetic Video to Real-World Benchmarks.} Despite training predominantly on synthetic videos, our approach demonstrates notable generalization to diverse real-world video understanding tasks. As shown in Tables~\ref{tab:main_results} and~\ref{tab:main_results2}, the model trained with SynRL achieves consistent improvements across multiple benchmark categories. On general video understanding tasks involving natural videos, we observe gains on VideoMMMU (particularly on the adaptation subset) and vinoground, indicating that temporal reasoning skills learned from synthetic data effectively transfer to complex real-world scenarios. 
The generalization is particularly evident on temporal grounding tasks with real-world videos. Table~\ref{tab:grounding_results} shows that training with our CoT-annotated synthetic videos, which provide frame-level timestamps and explicit temporal reasoning chains, substantially improves temporal localization capabilities across NExT-GQA, RexTime, and Charades-STA. These results confirm that the \textit{precise temporal annotations derived from our programmatic generation pipeline enable models to develop robust temporal grounding skills that transfer beyond synthetic scenarios to natural human activity videos.}

\noindent\textbf{Robust Task Transfer Across Grounding, Reasoning, and General Understanding.} Our approach exhibits strong cross-task generalization, with improvements spanning three distinct capability dimensions. For temporal grounding (Table~\ref{tab:grounding_results}), the model shows particularly strong gains on localization precision metrics across different IoU thresholds, indicating enhanced ability to precisely localize events in time. On temporal reasoning benchmarks (Tables~\ref{tab:main_results} and~\ref{tab:main_results2}), we observe improvements on TemporalBench and notable gains on complex reasoning tasks requiring cross-video understanding (TVbench, Minerva) and general video QA (MVBench, Aotbench). Importantly, these improvements in temporal capabilities do not compromise broader comprehension abilities, as evidenced by maintained or improved performance across diverse evaluation protocols.
This robust improvement pattern reveals a key insight: our synthetic data's temporal-centricity, where answers require holistic temporal integration rather than single-frame analysis, forces the model to develop fundamental temporal understanding that transfers across task boundaries. This uncovers an important finding: \textit{training on abstract motion patterns and simulated state transitions improves performance on both specialized temporal grounding and general video understanding.}

\begin{figure}[t!]
    \centering
    \includegraphics[width=1\linewidth]{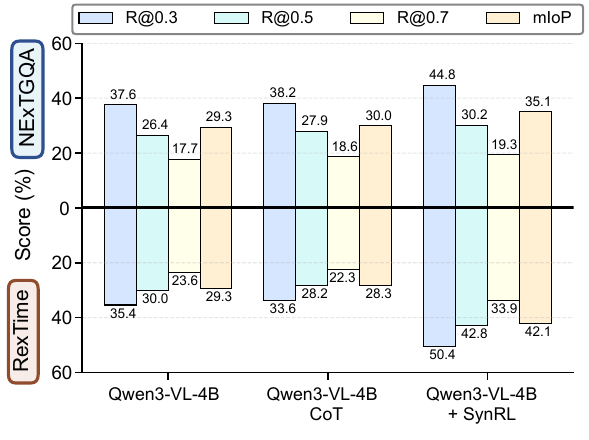}
    \caption{\textbf{mIoP Performance comparison on NExTGQA and RexTime benchmarks.}}
    \label{fig:miop}
\end{figure}

\section{Ablation Study and Analysis}

\noindent\textbf{Finding 1: Hard Training Samples Require Cold Start Before RL.} To validate the effectiveness of our two-stage training paradigm, we conduct ablation experiments comparing models trained with different strategies. As shown in Table~\ref{tab:ablation_coldstart}, incorporating a cold-start SFT phase with high-quality CoT data before RL training proves essential for achieving strong performance. Without cold-start initialization, the model struggles to learn from our challenging synthetic videos during RL, as the tasks require precise temporal reasoning that is difficult to bootstrap from sparse reward signals alone. The cold-start phase teaches the model how to perform step-by-step temporal reasoning, enabling more efficient exploration during subsequent RL training. This finding aligns with recent observations that RL on difficult tasks benefits from proper initialization—the cold-start phase provides a foundation of reasoning structure, while RL focuses on improving correctness and generalization.

\begin{table}[t]
\centering
\caption{Ablation study on the necessity of cold-start SFT before RL training. Models without cold-start fail to effectively learn from our challenging synthetic videos.}
\label{tab:ablation_coldstart}
\resizebox{0.5\textwidth}{!}{
\begin{tabular}{ccc|cccc}
\toprule
\multicolumn{3}{c|}{Training Strategy} & \multicolumn{4}{c}{Performance} \\
Base & SFT & RL & vinoground & TOMATO & MVBench & TVBench \\
\midrule
\checkmark & \ding{55} & \ding{55} & 40.8 & 32.1 & 65.4 & 51.9 \\
\checkmark & \ding{55} & \checkmark & 41.2 \uaa{0.4} & 32.8 \uaa{0.7} & 65.8 \uaa{0.4} & 52.1 \uaa{0.2} \\
\checkmark & \checkmark & \ding{55} & 42.6 \uaa{1.8} & 34.2 \uaa{2.1} & 66.1 \uaa{0.7} & 52.8 \uaa{0.9} \\
\checkmark & \checkmark & \checkmark & 43.4 \uaa{2.6} & 36.9 \uaa{4.8} & 66.8 \uaa{1.4} & 53.7 \uaa{1.8} \\
\bottomrule
\end{tabular}
}
\end{table}

\noindent\textbf{Finding 2: Quality Over Quantity, Synthetic Videos as Efficient Post-Training Data.} To demonstrate the advantages of our synthetic video approach, we compare cold-start initialization using Video-R1's CoT-165K dataset versus our 7.7K synthetic CoT data. As shown in Table~\ref{tab:quality_vs_quantity}, despite using over 20$\times$ more training samples, Video-R1's cold-start data leads to performance degradation on several benchmarks. Upon closer inspection of the Video-R1 CoT data, we identify systematic issues: the reasoning chains contain hallucinations about temporal events, incorrect motion descriptions, and logically inconsistent step-by-step analyses. When models are trained on such data, they fit these flawed reasoning patterns, leading to worse performance than the base model on tasks requiring precise temporal understanding.
In contrast, our programmatically generated synthetic videos with ground-truth annotations provide guaranteed correctness for both reasoning processes and final answers. This fundamental advantage—absolute correctness by construction—proves more valuable than scale. Training on 7.7K high-quality synthetic samples yields consistent improvements across benchmarks, demonstrating that for post-training temporal reasoning capabilities, data quality dramatically outweighs quantity. This finding has important implications: rather than scaling up potentially noisy model-generated annotations, focusing on smaller sets of verifiably correct training data offers a more reliable path to enhancing video understanding in VLMs.

\begin{table}[t]
\centering
\caption{Quality versus quantity in cold-start data. Our 7.7K synthetic CoT samples outperform Video-R1's 165K model-generated data across all benchmarks, demonstrating that correctness matters more than scale. TB. denotes TemporalBench.}
\label{tab:quality_vs_quantity}
\resizebox{0.5\textwidth}{!}{
\begin{tabular}{l|ccc}
\toprule
\textbf{Benchmark} & \textbf{ Base Model} & \textbf{Video-R1 CoT-165K} & \textbf{SynRL-7.7K (Ours)} \\
\midrule
vinoground & 40.8 & 36.6 \daa{4.2} & 42.6 \uaa{1.8} \\
TOMATO & 32.1 & 28.2 \daa{3.9} & 34.2 \uaa{2.1} \\
MVBench & 65.4 & 64.7 \daa{0.7} & 66.1 \uaa{0.7} \\
TVBench & 51.9 & 51.2 \daa{0.7} & 52.8 \uaa{0.9} \\
Charades & 41.9 & 41.4 \daa{0.5} & 43.4 \uaa{1.5} \\
TB. BA. & 67.4 & 63.7 \daa{3.7} & 67.9 \uaa{0.5} \\
TB. MBA. & 27.1 & 22.9 \daa{4.2} & 27.5 \uaa{0.4} \\
\bottomrule
\end{tabular}
}
\end{table}

\begin{figure}[t]
    \centering
    \includegraphics[width=0.99\linewidth]{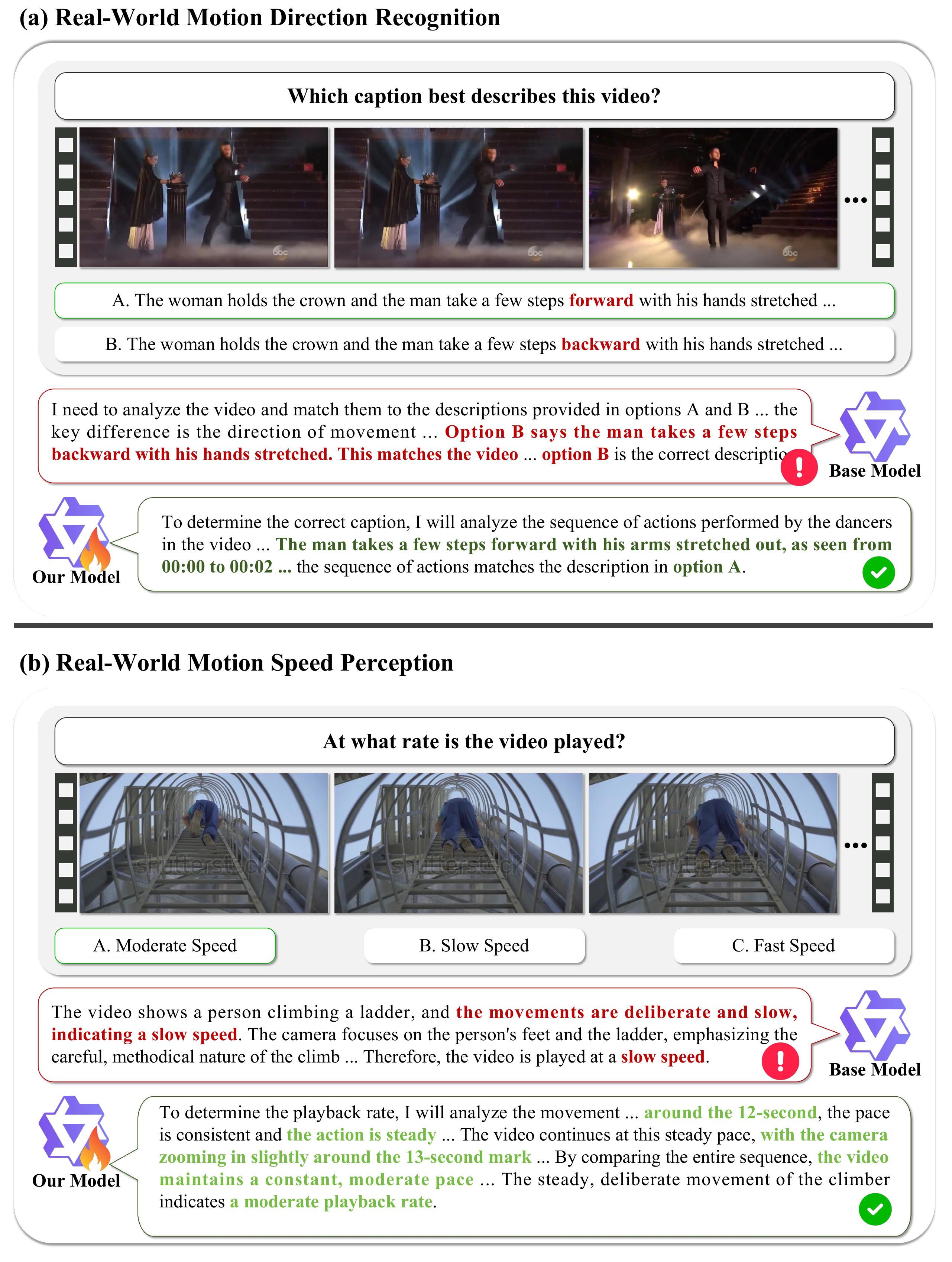}
    \caption{\textbf{Real-world motion understanding examples.}
    The base model fails on direction recognition and speed perception, while the SynRL-trained model answers both correctly, indicating that temporal skills acquired from synthetic videos can transfer to real-world video understanding.}
    \label{fig:case_study}
\end{figure}
\noindent\textbf{Finding 3: Efficient Scaling with Synthetic Videos.} To investigate the scalability of our approach, we explore training data scaling during the RL stage. Building upon evidence from prior work that demonstrates the positive correlation between RL data diversity/quantity and performance improvements, we leverage the convenient extensibility of synthetic data to scale along two novel dimensions: video duration and task complexity.
Specifically, we conduct a second round of RL training on top of our initial RL-trained model, focusing on more challenging long-term perception videos. We increase both the duration of generated videos and the number of intermediate state transitions. Since the first-round RL model has already established strong perceptual foundations, we construct 3K new RL samples with doubled video duration across all scenario types. We evaluate on representative benchmarks spanning different task categories.

\begin{table}[t!]
\centering
\caption{Performance improvements through iterative RL training with increasingly challenging synthetic videos.}
\label{tab:scaling}
\resizebox{0.5\textwidth}{!}{
\begin{tabular}{lcccc}
\toprule
\textbf{Model} & \textbf{Video-MME} & \textbf{AoTBench} & \textbf{Minerva} & \textbf{TOMATO} \\
\midrule
Base Model & 60.9 & 52.7 & 32.5 & 32.1 \\
+\,SynRL (1st round) & 62.0 \uaa{1.1} & 54.4 \uaa{1.7} & 32.8 \uaa{0.3} & 36.7 \uaa{4.6} \\
+\,+\,SynRL (2nd round) & 62.4 \uaa{0.4} & 54.9 \uaa{0.5} & 33.5 \uaa{0.7} & 37.0 \uaa{0.3} \\
\bottomrule
\end{tabular}
}
\end{table}

As shown in Table~\ref{tab:scaling}, the second round of RL training yields consistent performance gains across all benchmarks, demonstrating the effectiveness of our iterative scaling approach. These results highlight the efficiency and flexibility of our synthetic video generation method for RL scaling. This is particularly valuable given that high-quality, annotated long-form videos are extremely scarce in real-world datasets, making our programmatic generation approach a promising direction for future RL data scaling efforts.

\noindent\textbf{Finding 4: Emerging "Think with Time" Patterns from Synthetic to Real Videos.}
Our synthetic video training enables models to develop temporally grounded reasoning patterns that generalize effectively to real-world scenarios. As illustrated in Figure~\ref{fig:case_study}, we observe qualitative improvements in two fundamental temporal understanding capabilities. For motion direction recognition (Figure~\ref{fig:case_study}a), the base model incorrectly identifies the man's backward movement as forward motion, while the SynRL-trained model correctly analyzes "the sequence of actions" to determine the man moves backward. For speed perception (Figure~\ref{fig:case_study}b), the base model misclassifies moderate-speed climbing as slow motion based on superficial visual cues. In contrast, our model explicitly tracks temporal landmarks ("around the 12-second mark... around the 13-second mark") and compares "the entire sequence" to correctly identify the constant, moderate pace. These examples demonstrate that synthetic video training teaches models to perform frame-by-frame temporal analysis rather than relying on isolated visual patterns, enabling more robust real-world video understanding.

\section{Conclusion}
We introduce SynRL, a post-training framework that teaches temporal primitives through programmatically generated synthetic videos with guaranteed ground-truth annotations. Our key insight is that abstract temporal primitives—direction, speed, and state tracking—learned from simple geometric shapes transfer effectively to complex real-world scenarios. We construct high-quality CoT and RL samples by decomposing temporal understanding into short-term perceptual and long-term cognitive primitives, eliminating reliance on flawed proprietary model annotations.
Despite training on synthetic videos, SynRL achieves substantial improvements across 15 benchmarks and outperform Video-R1's 165K real-world samples. This establishes a new paradigm: systematic temporal learning through carefully designed synthetic data provides a more cost-efficient and reliable scaling path for video post-training.

\section*{Acknowledgement}

This work is supported by the National Key R\&D Program of China (Grant No. 2024YFC3308304), the "Pioneer" and "Leading Goose" R\&D Program of Zhejiang (Grant no. 2025C01128), and the ZJU-Angelalign R\&D Center for Intelligence Healthcare.

\section{Appendix}

\begin{figure}
    \centering
    \includegraphics[width=1\linewidth]{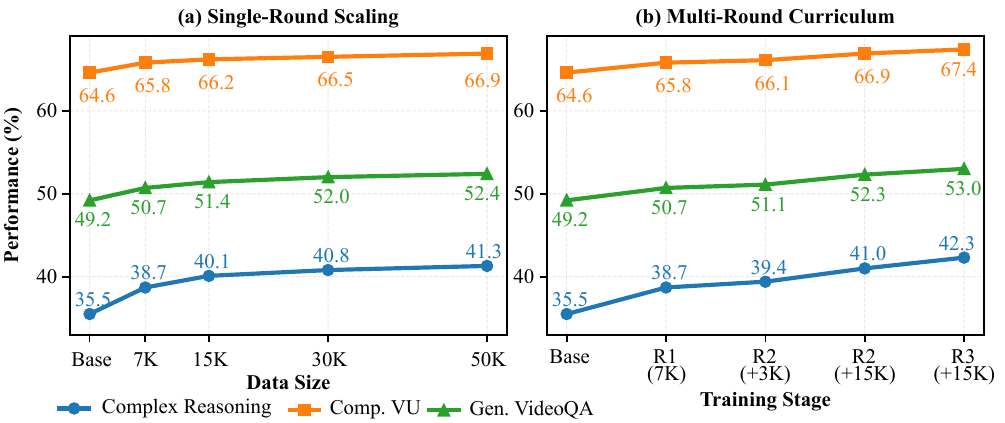}
    \caption{Scaling analysis}
    \label{fig:scaling}
\end{figure}

\begin{table}[t]
\centering
\caption{Comparison of synthetic vs.\ real-world training data across 
three evaluation categories. VideoR1 represents real-world data with 
model-generated annotations.}
\label{tab:syn_vs_real}
\resizebox{\columnwidth}{!}{%
\begin{tabular}{lccc}
\toprule
Model & Complex Reas. & Comp. VU & Gen. VideoQA \\
\midrule
Qwen3-VL-4B (Base) & 35.5 & 64.6 & 49.2 \\
\midrule
+ VideoR1 Only & 32.4 {\scriptsize\textcolor{red}{$-$3.1}} 
               & 62.7 {\scriptsize\textcolor{red}{$-$1.9}} 
               & 47.6 {\scriptsize\textcolor{red}{$-$1.6}} \\
+ Synthetic Only & \textbf{39.2} {\scriptsize\textcolor{blue}{+3.7}} 
                 & 65.6 {\scriptsize\textcolor{blue}{+1.0}} 
                 & 49.9 {\scriptsize\textcolor{blue}{+0.7}} \\
+ Mixed          & 38.7 {\scriptsize\textcolor{blue}{+3.2}} 
                 & \textbf{65.8} {\scriptsize\textcolor{blue}{+1.2}} 
                 & \textbf{50.7} {\scriptsize\textcolor{blue}{+1.5}} \\
\bottomrule
\end{tabular}%
}
\end{table}

\begin{table*}[ht]
\centering
\caption{Per-category ablation study showing capability-specific 
transfer patterns of each synthetic primitive type. 
TOM.\ = TOMATO; MV.\ = MVBench; VE.\ = VideoMME; 
TB = TemporalBench; VSI = VSIBench.
Best results in each column are in \textbf{bold}.}
\label{tab:per_category}
\resizebox{0.85\textwidth}{!}{%
\begin{tabular}{lcccccccc}
\toprule
Training Data & TOM. & MV.-action & MV.-move 
              & VE.-short & VE.-long 
              & TB-action & TB-motion & VSI \\
\midrule
Base Model 
  & 32.1 & 66.3 & 70.1 & 71.7 & 51.1 & 63.1 & 64.2 & 35.0 \\
\midrule
+ Spatial 
  & 33.8 {\scriptsize\textcolor{gray}{+1.7}} 
  & 67.1 {\scriptsize\textcolor{gray}{+0.8}} 
  & \textbf{74.3} {\scriptsize\textcolor{blue}{+4.2}} 
  & 72.3 {\scriptsize\textcolor{gray}{+0.6}} 
  & 51.8 {\scriptsize\textcolor{gray}{+0.7}} 
  & 64.2 {\scriptsize\textcolor{gray}{+1.1}} 
  & 65.1 {\scriptsize\textcolor{gray}{+0.9}} 
  & \textbf{38.4} {\scriptsize\textcolor{blue}{+3.4}} \\
+ Temporal 
  & 35.9 {\scriptsize\textcolor{blue}{+3.8}} 
  & 68.7 {\scriptsize\textcolor{blue}{+2.4}} 
  & 71.2 {\scriptsize\textcolor{gray}{+1.1}} 
  & 72.1 {\scriptsize\textcolor{gray}{+0.4}} 
  & 53.6 {\scriptsize\textcolor{blue}{+2.5}} 
  & \textbf{67.4} {\scriptsize\textcolor{blue}{+4.3}} 
  & 65.8 {\scriptsize\textcolor{gray}{+1.6}} 
  & 35.9 {\scriptsize\textcolor{gray}{+0.9}} \\
+ Short-term 
  & 33.2 {\scriptsize\textcolor{gray}{+1.1}} 
  & 67.8 {\scriptsize\textcolor{gray}{+1.5}} 
  & 72.5 {\scriptsize\textcolor{gray}{+2.4}} 
  & 73.1 {\scriptsize\textcolor{gray}{+1.4}} 
  & 51.5 {\scriptsize\textcolor{gray}{+0.4}} 
  & 65.6 {\scriptsize\textcolor{gray}{+2.5}} 
  & 67.2 {\scriptsize\textcolor{blue}{+3.0}} 
  & 36.7 {\scriptsize\textcolor{gray}{+1.7}} \\
+ Long-term 
  & 36.2 {\scriptsize\textcolor{blue}{+4.1}} 
  & 67.2 {\scriptsize\textcolor{gray}{+0.9}} 
  & 70.8 {\scriptsize\textcolor{gray}{+0.7}} 
  & 71.9 {\scriptsize\textcolor{gray}{+0.2}} 
  & 54.1 {\scriptsize\textcolor{blue}{+3.0}} 
  & 64.8 {\scriptsize\textcolor{gray}{+1.7}} 
  & 64.5 {\scriptsize\textcolor{gray}{+0.3}} 
  & 36.1 {\scriptsize\textcolor{gray}{+1.1}} \\
\midrule
+ Full SynRL 
  & \textbf{36.7} {\scriptsize\textcolor{blue}{+4.6}} 
  & \textbf{69.1} {\scriptsize\textcolor{blue}{+2.8}} 
  & 73.8 {\scriptsize\textcolor{blue}{+3.7}} 
  & \textbf{73.4} {\scriptsize\textcolor{blue}{+1.7}} 
  & \textbf{54.5} {\scriptsize\textcolor{blue}{+3.4}} 
  & 67.1 {\scriptsize\textcolor{blue}{+4.0}} 
  & \textbf{67.8} {\scriptsize\textcolor{blue}{+3.6}} 
  & 38.2 {\scriptsize\textcolor{blue}{+3.2}} \\
\bottomrule
\end{tabular}%
}
\end{table*}
\subsection{Related Work}

\noindent\textbf{Reinforcement Learning for Video Understanding.} 
In recent years, vision-language models (VLMs) have achieved remarkable 
progress across diverse tasks through supervised fine-tuning (SFT)-based 
training paradigms~\cite{li2025frequency,li2025sepprune,li2025preference,li2025efficient,li2025mmt,li2025ammkd,li2024sglp,li2026comprehensive,DBLP:journals/corr/abs-2509-18600,DBLP:conf/nips/LimC0B0HL24,li2025retidiff,li2026cross,li2026garnet,hou2025codev,zhao2025towards,zhang2025cross,Guan_2025_ICCV,10887029,wei2026univbenchunifiedevaluationvideo,jiang2024med,jiang2025hulu,pan2025dusss,pan2026frequency,dou2025plan,zhu2025pathology}. Building upon this foundation, and inspired by the 
success of Reinforcement Learning from Verifiable Rewards (RLVR) in 
language models~\cite{lambert2024tulu3,chen-etal-2025-datasets,luo2026narrowpanoramicvisionattentionguided,luounlocking,zhu2025medeyes}, recent work 
has begun adapting this paradigm to video understanding tasks. Video-R1~\cite{feng2025videor1reinforcingvideoreasoning} introduces Temporal Group Relative Policy Optimization (T-GRPO), which explicitly encourages temporal modeling by training on both correctly ordered and randomly shuffled frame sequences, achieving substantial improvements with only 165K training samples. VideoChat-R1~\cite{li2025videochatr1enhancingspatiotemporalperception} demonstrates that reinforcement fine-tuning is highly data-efficient for spatio-temporal perception tasks, achieving promising improvement in temporal grounding. TinyLLaVA-Video-R1~\cite{zhang2025tinyllavavideor1} shows that models under 4B parameters can develop sophisticated video reasoning through RL, exhibiting emergent "aha moments" of self-correction. More recent work explores pure RL training without supervised fine-tuning~\cite{wang2025videorts}, test-time scaling mechanisms~\cite{wang2025videorts}, and specialized applications including temporal grounding~\cite{chen-etal-2025-datasets}, tool-augmented reasoning~\cite{zhang2025vital}, and video anomaly understanding~\cite{chen2025vaur1}. These approaches consistently demonstrate that RL outperforms supervised fine-tuning for video temporal reasoning, achieving improvements while using less training data.

\subsection{Further Analysis}
\noindent\textbf{Scaling Analysis:}
We conducted scaling experiments along two dimensions: (1) single-round data scaling with increasing dataset sizes, and (2) iterative multi-round training with progressively challenging videos to test curriculum-based scalability.

\noindent\textbf{Key Findings:} (1) \textbf{No saturation observed:} Performance improves consistently up to 50K samples across all categories, with complex reasoning showing strongest scaling. (2) \textbf{Iterative training achieves higher ceiling:} Three-round curriculum training (37K total) yields higher gains than single-round 50K training, demonstrating that progressive difficulty increase is more effective than simply scaling data volume. (3) 7k-10k is the best size for single RL turn.

\noindent\textbf{CoT Analysis}
We conducted a \textbf{blind evaluation} on 50 examples using Gemini-3-Pro and 4 human reviewers (each assessing 12 examples). The SynRL model outperformed the base model with an \textbf{82\% win-rate}, demonstrating three key improvements: explicit timestamped reasoning, active state tracking, and stronger reasoning-answer consistency.

\noindent\textbf{Synthetic vs.\ Real-World Data.}
Table~\ref{tab:syn_vs_real} compares three training data configurations.
First, real-world data with noisy model-generated annotations (VideoR1)
consistently degrades performance, indicating that the current bottleneck
lies in temporal reasoning rather than scene understanding.
Second, synthetic-only training outperforms VideoR1, demonstrating that
abstract temporal primitives transfer effectively to diverse real-world
scenarios, as base models already possess sufficient visual understanding
capabilities.
Third, mixed training achieves the best overall performance by combining
synthetic data's precise temporal reasoning with real-world data's visual
diversity, facilitating broader generalization.

\noindent\textbf{Per-Category Ablation.}
To further understand the contribution of each primitive type,
we decompose our synthetic data along two orthogonal dimensions:
(1) \emph{Spatial} primitives (direction, speed, trajectory) vs.\
\emph{Temporal} primitives (state tracking, symbol operations);
(2) \emph{Short-term} vs.\ \emph{Long-term} videos.
Table~\ref{tab:per_category} reports results on representative
benchmark subsets targeting specific capabilities.

Spatial primitives transfer most effectively to spatial reasoning
benchmarks (VSIBench, MVBench-moving), while temporal primitives
dominate action understanding and complex reasoning
(TB-action, TOMATO).
Short-term training benefits immediate perception tasks
(TB-motion, VideoMME-short), whereas long-term training is
critical for temporal and long-video reasoning
(TOMATO, VideoMME-long).
The full SynRL model, combining all primitive types, achieves
consistent improvements across all categories, confirming that
the four primitive types capture complementary reasoning skills.

\noindent\textbf{Pass@$k$ Analysis.}
To assess the model's reasoning potential beyond single-sample
performance, we evaluate Pass@$k$ with $k\in\{1,4\}$ by sampling
four independent responses per question and reporting the fraction
of questions answered correctly at least once ($k=4$).
As shown in Table~\ref{tab:passk}, SynRL exhibits a larger
improvement margin from Pass@1 to Pass@4
(+3.9 average) compared to the base model (+2.1 average),
with the most pronounced gain on Complex Reasoning tasks.
This indicates that learned temporal primitives enable the model
to explore correct reasoning directions more reliably,
reflecting a higher performance ceiling rather than mere
exploitation of surface-level patterns.

\begin{table}[t]
\centering
\caption{Pass@$k$ performance comparison ($k\in\{1,4\}$).
SynRL shows a larger gap between Pass@1 and Pass@4,
indicating a higher reasoning potential ceiling.}
\label{tab:passk}
\resizebox{\columnwidth}{!}{%
\begin{tabular}{lcccc|ccc}
\toprule
\multirow{2}{*}{Model} 
  & \multicolumn{4}{c|}{Pass@1} 
  & \multicolumn{3}{c}{Pass@4} \\
\cmidrule{2-8}
  & Complex & Comp.\ VU & Gen.\ VQA & VideoMME
  & Complex & Comp.\ VU & Gen.\ VQA \\
\midrule
Base Model 
  & 35.5 & 64.6 & 49.2 & 58.2
  & 37.8 & 66.3 & 51.5 \\
\midrule
+ SynRL 
  & 38.7 {\scriptsize\textcolor{blue}{+3.2}} 
  & 65.8 {\scriptsize\textcolor{blue}{+1.2}} 
  & 50.7 {\scriptsize\textcolor{blue}{+1.5}} 
  & 60.5 {\scriptsize\textcolor{blue}{+2.3}}
  & \textbf{42.6} {\scriptsize\textcolor{blue}{+4.8}} 
  & \textbf{68.4} {\scriptsize\textcolor{blue}{+3.1}} 
  & \textbf{53.4} {\scriptsize\textcolor{blue}{+1.9}} \\
\bottomrule
\end{tabular}%
}
\end{table}
\noindent\textbf{Effect of Chain-of-Thought Supervision.}
Our pipeline provides code-generated metadata as accurate reasoning
traces, serving as high-quality cold-start supervision for learning
temporal primitives before RL fine-tuning.
To quantify this contribution, Table~\ref{tab:cot_ablation}
compares three configurations: (1) CoT warm-up followed by RL
training on 7K samples (\emph{CoT + RL}); (2) RL training only
on 7K samples; and (3) RL training only on 15K samples.

RL-only training at 7K achieves modest but consistent gains,
confirming that synthetic videos are beneficial even without
explicit CoT supervision. Scaling RL-only to 15K largely closes
the gap with CoT + RL at 7K, indicating that CoT supervision
primarily accelerates learning efficiency rather than unlocking
qualitatively different capabilities.
Nonetheless, CoT + RL at 7K remains competitive with RL-only
at 15K while using fewer samples, demonstrating that high-quality
reasoning traces provide an effective data-efficient shortcut
for acquiring temporal primitive skills.

\begin{table}[t]
\centering
\caption{Ablation study on Chain-of-Thought (CoT) supervision.
CoT warm-up accelerates RL learning but models still benefit
from synthetic videos without CoT supervision when given
more data.}
\label{tab:cot_ablation}
\resizebox{\columnwidth}{!}{%
\begin{tabular}{lccc}
\toprule
Model & Complex Reas. & Comp. VU & Gen. VideoQA \\
\midrule
Qwen3-VL-4B (Base) & 35.5 & 64.6 & 49.2 \\
\midrule
+ CoT + RL (7K)  
  & 38.7 {\scriptsize\textcolor{blue}{+3.2}} 
  & \textbf{65.8} {\scriptsize\textcolor{blue}{+1.2}} 
  & 50.7 {\scriptsize\textcolor{blue}{+1.5}} \\
+ Only RL (7K)   
  & 37.1 {\scriptsize\textcolor{blue}{+1.6}} 
  & 65.2 {\scriptsize\textcolor{blue}{+0.6}} 
  & 50.0 {\scriptsize\textcolor{blue}{+0.8}} \\
+ Only RL (15K)  
  & \textbf{39.0} {\scriptsize\textcolor{blue}{+3.5}} 
  & 65.6 {\scriptsize\textcolor{blue}{+1.0}} 
  & \textbf{50.9} {\scriptsize\textcolor{blue}{+1.7}} \\
\bottomrule
\end{tabular}%
}
\end{table}



{
    \small
    \bibliographystyle{ieeenat_fullname}
    \bibliography{main}
}

\end{document}